# Evaluation of handwriting kinematics and pressure for differential diagnosis of Parkinson's disease

Peter DROTÁR[a], Jiří MEKYSKA[a], Irena REKTOROVÁ[b], Lucia MASAROVÁ[b], Zdeněk SMÉKAL[a], Marcos FAUNDEZ-ZANUY[c]

[a]Department of Telecommunications, Brno University of Technology, Technická 12, 61200 Brno, Czech Republic

[b] First Department of Neurology, Faculty of Medicine, St. Anns University Hospital, Pekarska 664, 66591 Brno, Czech Republic

[c] Signal Processing Group, Tecnocampus, Escola Universitaria Politecnica de Mataro, Avda. Ernest Llunch 32, 08302, Mataro, Spain.

## Abstract

*Objective:* We present the PaHaW Parkinson's disease handwriting database, consisting of handwriting samples from Parkinson's disease (PD) patients and healthy controls. Our goal is to show that kinematic features and pressure features in handwriting can be used for the differential diagnosis of PD.

*Methods and Material:* The database contains records from 37 PD patients and 38 healthy controls performing eight different handwriting tasks. The tasks include drawing an Archimedean spiral, repetitively writing orthographically simple syllables and words, and writing of a sentence. In addition to the conventional kinematic features related to the dynamics of handwriting, we investigated new pressure features based on the pressure exerted on the writing surface. To discriminate between PD patients and healthy subjects, three different classifiers were compared: K-nearest neighbors (K-NN), ensemble AdaBoost classifier, and support vector machines (SVM).

*Results:* For predicting PD based on kinematic and pressure features of handwriting, the best performing model was SVM with classification accuracy of $P_{acc}$ = 81.3% (sensitivity $P_{sen}$ = 87.4% and specificity of $P_{spe}$ = 80.9%). When evaluated separately, pressure features proved to be relevant for PD diagnosis, yielding $P_{acc}$ = 82.5% compared to $P_{acc}$ = 75.4% using kinematic features.

*Conclusion:* Experimental results showed that an analysis of kinematic and pressure features during handwriting can help assess subtle characteris-tics of handwriting and discriminate between PD patients and healthy controls.

## 1. Introduction

Parkinson's disease is a complex neurodegenerative disease that affects a large portion of the worldwide population [1]. With current prevalence rates, ranging from 10 to 800 people per 100,000, PD is one of the most common neurodegenerative disorders [2]. PD is a movement disorder characterized by resting tremor, rigidity, slowness of movement (bradykinesia), and loss of postural reflexes. The disturbances of motor control in PD involve processing of motor planning, motor programming, motor sequencing, movement initiation and movement execution [3].

There is currently no objective method for diagnosing PD. It can take months to get a reliable PD diagnosis, and symptoms need to be carefully monitored. Even then the probability of an inaccurate diagnosis is approximately 25 % [4]. The diagnosis can be confirmed only by a pathological analysis at autopsy; this further highlights the complexity of the diagnosis. Decision support tools for accurate diagnosis would be beneficial for early diagnosis and for the development of treatment strategies for PD patients [5, 6]. Identifying biomarkers is an important goal of the research on neurodegenerative diseases [7].

One typical hallmark of PD is disruption in the execution of practiced skills such as handwriting [8, 9]. People with PD frequently have severe difficulties in coordinating of the components of a motor sequence movement. They tend to perform sequential movements in a more segmented fashion. Hesitations and pauses are often observed between the components of the sequence [10, 11]. Continuous handwriting and similar motor tasks occur more slowly than in a healthy person. Some recent studies have suggested that handwriting can be used as a biomarker for diagnosing PD [12, 13]. The reasoning behind this suggestion is that handwriting is no longer an automated process for PD patients and their handwriting depends on a visual closed loop [14].

Several handwriting tasks were proposed for use in analyzing the handwriting of PD patients and for obtaining insight into the motor disruption caused by PD. Probably the most popular handwriting exercise for tremor assessment is currently the Archimedean spiral. Spiral drawing has been frequently used for evaluating of the motor performance in various movement disorders, including PD [4, 8, 15–17]. Words containing one or more repetitions of the cursive letter "l" are the second-most common exercises in handwriting assessment [10, 18]. In addition to these established tasks, we proposed new ones consisting of writing simple words and short sentences. The words used in these handwriting tasks were selected for their simple orthography and easy syntax.

It has been shown that the absolute positioning of the pen during handwriting is relevant for PD diagnosis, as are pen movements above the writing surface (when the pen does not leave the trajectory) [12, 13, 19]. Pressure exerted on the surface during handwriting plays a significant role too [13].

In this paper, we extend our previous work [12, 20] by providing a more detailed analysis of the pressure modality of handwriting and by introducing novel pressure features. Moreover, we introduce the Parkinson's disease handwriting (*PaHaW* ) database, which can be used for developing predictive models for PD diagnosis. The *PaHaW* database contains recorded in-air/on-surface trajectories and pressure, i.e. modalities that have been shown to be significant for PD classification. The results confirmed that handwriting is relevant in diagnosing and monitoring PD. We also compared three frequently used classifiers on the *PaHaW* database: SVM, Adaboost and K-NN.

We believe that the *PaHaW* database can encourage further research and provide additional information to other available databases related to PD such as Parkinson's disease speech datasets [21, 22].

In the next section, the database of handwriting samples is introduced and described in detail. The third section presents our methods and obtained results. We provide a discussion and conclusions in the last section.

## 2. Parkinson's disease handwriting (PaHaW) database

We created a handwriting database of 37 PD patients (19 men) and 38 sex- and age- matched healthy controls (20 men). The database was acquired in cooperation with the First Department of Neurology, Masaryk University and St. Anne's University Hospital in Brno, Czech Republic.

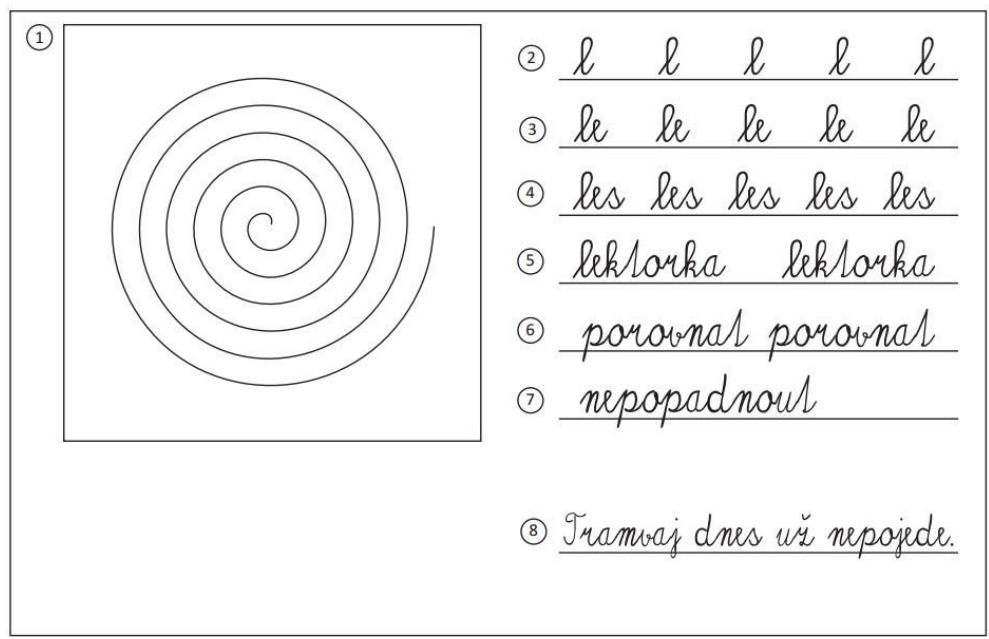

Figure 1: Illustration of filled template (not actual handwriting samples)

Subjects were rested and seated in front of the table in comfortable position. Each subject was asked to complete a handwriting task according to the prepared pre-filled template at a comfortable speed. Subjects were allowed to repeat the task in case of some error or mistake during handwriting. The pre-filled template is depicted in Fig. 1 [23]. The pre-filled template was shown to the subjects; no restrictions about the number of repetitions of syllables/words in tasks or their height were given.

A tablet was overlaid with an empty paper template (containing only printed lines and square box specifying area for Archimedean spiral), and a special ink pen was held in a normal fashion, allowing for immediate full visual feedback. The signals were recorded using the Intuos 4M (Wacom technology) digitizing tablet.

Digitized signals were acquired during the movements executed while exerting pressure on the writing surface (on-surface movement) and during the movement above the writing surface (in-air movement). The perpendicular pressure exerted on the tablet surface was also recorded. The recordings started when the pen touched the surface of the digitizer and finished when the task was completed. The tablet captured the following dynamic features (time-sequences): x-coordinate, $x[t]$; y-coordinate, $y[t]$; time stamp, $s[t]$; button status, $b[t]$; pressure, $p[t]$; and discrete time $t$. Button status is a binary variable, being 0 for in-air movement and 1 for on-surface movement.

The tablet sampling rate was 100 samples per second; the acquisition software was developed by the research team. Subsequent analysis was performed using Matlab and Python programming language.

## 2.1. Subjects

Altogether, 75 subjects (37 PD patients and 38 healthy controls) participated in the study. The participants were enrolled in the First Department of Neurology, St. Anne's University Hospital in Brno. A complete list of all participants is provided in Appendix A, with information about sex, age, disease duration, UPDRS-part V score[1] and levodopa equivalent daily dose[2]. Mean and standard deviation of age, UPDRS-Part V - Modified Hoehn and Yahr staging score [24] and disease duration are summarized in Table 1. No significant differences related to gender or age were found between the PD and healthy control groups.

All the subjects had completed at least ten years of education and reported Czech as their native language. All subjects used their dominant right hand. None of the subjects had a history or presence of any psychiatric symptoms or any disease affecting the central nervous system (other than PD in the PD cohort). All PD patients completed the tasks under L-DOPA medication.

Prior to handwriting acquisition, each patient was evaluated by a clinical neurologist. The healthy controls were examined by a clinician in order to make sure that there was no movement disorder or injury present that could significantly affect handwriting. We removed subjects whose handwriting was apparently affected by another diseases or who were not in the suitable physical condition. Basic information and instructions regarding the upcoming task were provided for each subject, and they were allowed to practice the task before the recording.

---

[1]Unified Parkinson's Disease Rating Scale (UPDRS) is a rating scale used to follow the longitudinal course of PD [24]

[2]levodopa equivalent dose (LED) of a drug that produces the same anti-parkinsonian effect as 100 mg of immediate-release levodopa

Table 1: Parkinson's handwriting dataset. Characteristics of healthy controls (H) and Parkinson's disease (PD) group.

|  | age | | UPDRS (part V) | | years since diag. | | LED | | male/female |
|---|---|---|---|---|---|---|---|---|---|
|  | *mean* | *std* | *mean* | *std* | *mean* | *std* | *mean* | *std* |  |
| PD | 69.3 | 10.9 | 2.27 | 0.84 | 8.37 | 4.8 | 1373.4 | 714 | 19/18 |
| H | 62.4 | 11.3 | - | - | - | - | - | - | 20/18 |

## 2.2. Handwriting tasks

The template consisted of eight different handwriting tasks. Based on a survey of the literature, we included drawing of an Archimedean spiral and repetitively writing cursive a letter "l", or a syllable "le", respectively.

The Archimedean spiral is an established task used in assessing of akinesia in PD and essential tremor [16, 18]. During this task, the template was shown to the subject for visual guidance. Subjects drew the spiral from inside to out, but were not asked to draw spiral within particular boundaries or to follow a pre-drawn line.

In tasks 2, 3, and 4 participants wrote the cursive letter l, bigram le or the trigram les. Similar tasks (writing the letter l – or its variations) are frequently used for handwriting analysis [10, 17].

Tasks 5, 6, and 7 were to write words lektorka – female teacher, porovnat – to compare, and nepopadnout - to not catch (written in Czech –, the native language of all participants). These words are characterized by simple orthography and quite easy syntax. The common characteristic is that they can be written continuously, without lifting the pen above the surface, i.e. they can be written in one continuous movement.

Task 8 was to write a longer sentence: Tramvaj dnes už nepojede (The tram won't go today). Use of the whole sentence allowed us to acquire also movements above the writing surface, i.e. in-air movements, during transitions between individual words in the sentence.

## 3. Methods and results

### 3.1. Feature extraction

The handwriting features were computed from on-surface movements (in the form of Cartesian coordinates) and pressure. The kinematic features used in this study are listed in Table 2 [12]. The term stroke represents single connected continuous component of trace, i.e. on-surface movement between two successive pen lifts. According to this definition spiral or letter l are usually drawn as one stroke. Strokes were used only to calculate stroke speed.

Table 2: Overview of kinematic handwriting features

| Feature | Description |
|---|---|
| stroke speed | stroke length divided by stroke duration in $mm/s$ |
| speed | trajectory during handwriting divided by handwriting duration in $mm/s$ |
| velocity | rate at which the position of a pen changes with time in $mm/s$ |
| acceleration | rate at which the velocity of a pen changes with time in $mm/s^2$ |
| jerk | rate at which the acceleration of a pen changes with time in $mm/s^3$ |
| horizontal velocity/acceleration/jerk | velocity/acceleration/jerk in horizontal direction |
| vertical velocity/acceleration/jerk | velocity/acceleration/jerk in vertical direction |
| number of changes in velocity direction (NCV) | the mean number of local extrema of velocity ([25]) |
| number of changes in acceleration direction (NCA) | the mean number of local extrema of acceleration ([25]) |
| relative NCV | NCV relative to writing duration |
| relative NCA | NCA relative to writing duration |
| on-surface time | time spent on-surface during writing |
| normalised on-surface time | time spent on-surface during writing normalised by whole writing duration |

Novel pressure handwriting features were computed to take advantage of all tablet functionalities. The typical pressure profile during writing is depicted in Fig. 2 for Archimedean spiral and Fig. 3 for tasks 2, 4 and 6.

The fundamental pressure features are the value of *pressure* as captured by the tablet during the particular task and the rate at which *pressure changes with respect to time*. Similarly to the concept of the number of changes in velocity [25] we proposed a *number of changes in pressure* (NCP) and *relative NCP*. Since there can be rapid changes in pressure that lead to incorrect NCP, we smoothed the data using a local regression using weighted linear least squares and a first degree polynomial model[3]. Relative NCP is NCP normalised by the whole length of writing. We introduced correlation coefficients to capture the relationship between pressure and kinematic features. In particular, we computed *correlation between pressure and (horizontal/vertical) velocity/acceleration*. Altogether, six correlation coefficients $\rho_{(horizontal/vertical),vel/acc}$ were computed.

---
[3]We used Matlab's built-in function *smooth*

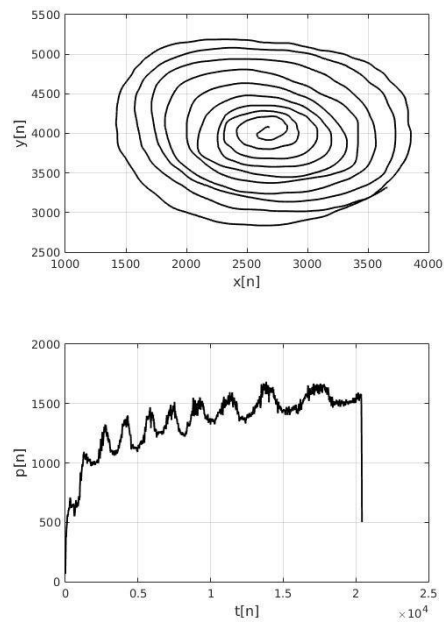

Figure 2: Example of handwriting from task 1 (Archimedean spiral). On surface trajectory recorded by tablet (top figure) and related pressure exerted on tablet surface (bottom figure).

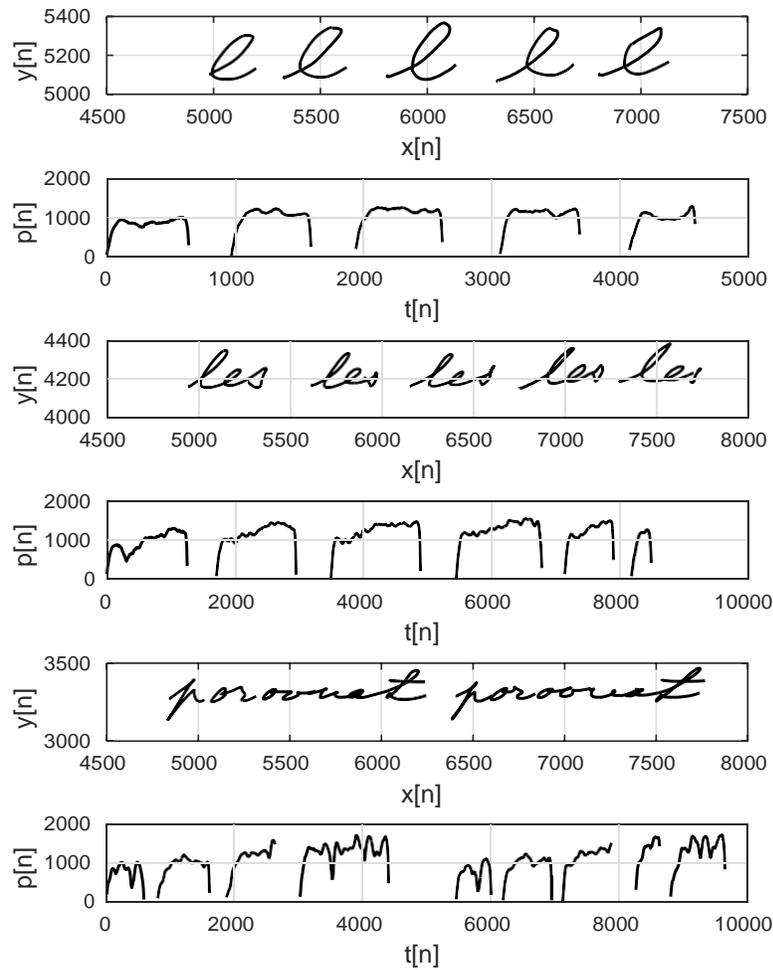

Figure 3: Example of handwriting from tasks 2, 4 and 6. On surface trajectory recorded by tablet and related pressure exerted on tablet surface.

Fig. 2 and Fig. 3 shows that the main trend of the typical pressure trajectory starts with a rising edge, continues with a slowly increasing main movement, and finishes with a falling edge. The progress of the main trend is similar for all tasks. On the other hand, there is visible difference in smoothness of the main part of the signal for different tasks. As can be seen from the Fig. 3, the main part is relatively smooth for the second task, however, as the performed task become longer the main part is more rough. The exterted pressure varies with the length of the word and complexity of the drawn letter. This may indicate why are some tasks more useful for classification than others.

The drawing of the spiral is different from the other tasks used in our study. Handwriting features are generally more stable for writing of the words, whereas, when drawing the spiral features change from beginning to the end of the task. This can be seen in Fig. 2 where the pressure trajectory is continuously increasing. In this study, we focus on the analysis of the handwriting and we did not introduce any spiral specific features. Therefore we extracted the same features from signals from all tasks.

We expected that there would be indicators for particular parts of the pressure signal, therefore we calculated the above mentioned *pressure features* separately for rising edge, main movement, and falling edge. The features corresponding to the different parts of the signal are denoted by superscript *rise*, *main* or *fall*. The boundary between the edges and the main movement is given by a median of signal pressure. Additionally, the range (maximal value - minimal value) between the rising edge and falling edge time duration ($R_{time}^{fall/rise}$) and the range of the rising edge and falling edge pressure ($R_{press}^{fall/rise}$) were included in the analysis. The final included feature was the *pressure overshoot*, providing the distance between the pressure maximum and pressure median.

Additionally, six basic statistical functionals (mean, median, standard deviation, 1st percentile, 99th percentile, 99th percentile - 1st percentile) were computed. Features were normalised before classification on a per-feature basis to have a zero mean and a standard deviation of one.

### 3.2. Statistical classification

Our aim was to build a discriminative model to differentiate between people with PD and healthy subjects. It is a binary classification task that can be resolved by statistical machine learning algorithms.

We compared three frequently used machine learning techniques: SVM [26], AdaBoost classifier with a decision tree base estimator [27] and $K$-NN algorithm. We used *Python* implementation of *scikit-learn* library [28].

The underlying idea of SVM classifiers is to calculate a maximal margin hyperplane separating two classes of the data. To learn non-linearly separable functions, the data are implicitly mapped to a higher dimensional space by means of a kernel function, where a separating hyperplane is found. New samples are classified according to the side of the hyperplane they belong to. We used radial basis function (RBF) kernel [26]. The RBF kernel is defined as

$$K(x, x_i) = e^{\frac{-\|x - x_i\|^2}{2\gamma^2}} \tag{1}$$

where $\gamma$ controls the width of RBF function.

The parameters kernel gamma $\gamma$ and penalty parameter $C$ were optimized using grid search of possible values. Specifically, we searched over the grid $(C, \gamma)$ defined by the product of the sets $C = [2^{-8}, 2^{-5}, \ldots, 2^7, 2^8]$, $\gamma = [2^{-9}, 2^{-4}, \ldots, 2^8, 2^9]$.

AdaBoost is one of the important *ensemble methods* known as *boosting*. The key idea behind boosting techniques is to use ensemble methods to combine weak classifiers in order to build a strong learner. AdaBoost is an iterative boosting algorithm constructing a strong classifier as a linear combination of weak classifiers, each performing at least above chance level (50 % correct classification). We used decision trees classifiers as weak classifiers [29]. The maximum number of estimators at which boosting is terminated was set to 500. Settings used for decision trees were as follows. The number of features to consider when looking for the best split was the square root of the number of features and the maximum depth of the tree was set to 3.

In the $K$-NN algorithm, $k$-the nearest samples in a reference set are found, by taking a majority vote among the classes of these $k$ samples. The goal is to determine the true class of an undefined test pattern by finding the nearest neighbors within a hyper-sphere of predefined radius. For the $K$-NN classifier, the best results were obtained with $k = 3$.

### 3.3. Numerical results

Classifier validation was conducted using stratified 10-fold cross-validation. The data was divided into ten mutually exclusive and exhaustive equal-sized subsets. For each subset, the union of all other subsets was considered as training data and the error rate was determined. Errors over different subsets were averaged to obtain the classification error. The process was repeated a total of ten times; the original dataset was randomly permuted in each repetition prior to splitting into training and testing subsets. The results were averaged over all ten runs.

The classification test performance was determined by the classification accuracy $P_{acc}$, sensitivity $P_{sen}$ and specificity $P_{spe}$ [30].

From all computed features we kept only those that passed the Mann-Whitney U test, i.e. those that showed a statistically significant ($p < 0.05$) difference between the PD and control groups. Table 3 shows 20 most relevant features and median of their values for PD and healthy control group. Features are sorted according Spearman's correlation coefficient $\rho$.

At first, we evaluated the prediction potential of different handwriting tasks considering both conventional kinematic features and pressure features. The classification accuracies for all tasks are depicted in Table 4. We did not find any statistically significant kinematic features for the tasks 1 and 4; therefore, for these two tasks we considered only pressure features. The highest classification accuracy using pressure features $P_{acc}$ = 74.2% was obtained for data from the task 6. The slightly lower classification accuracy of 73.2% was achieved for the task 8. The best results for kinematic features were provided by data from the task 8. Similarly, task 8 was the most discriminative for merged kinematic and pressure features. Both modalities, i.e. pressure and kinematic features showed relatively similar classification accuracies for all tasks. Interestingly, the fusion of pressure and kinematic features[3] did not result in any improvement in terms of classification accuracy. The only exception was the most predictive task 8.

Merging all of the tasks together noticeably improved the classification accuracy, probably due to the different nature of the handwriting tasks. The results presented in Table 4 were obtained using an SVM classifier. To obtain more confidence in our results and to compare different classifiers we also employed the AdaBoost classifier and $K$-NN classifier. Overall classification accuracy ($P_{acc}$), sensitivity ($P_{sen}$), and specificity ($P_{spe}$) for all tasks and merged kinematic and pressure features are provided in Table 5. Comparing all three classifiers, it is clear that the best results in terms of accuracy, specificity, and sensitivity were obtained using the SVM classifier.

---

[3]Only tasks 2 to 8 were merged. Task 1 contained data from only 69 subjects and did not show any significant discrimination potential, therefore we did not include this task.

Table 3: Twenty kinematic and pressure features with largest relevance to class label sorted according to the Spearman's correlation coefficient $|\rho|$. Displayed are medians and standard deviations (std) for healthy controls (H) and Parkinson's disease group.

| feature, stat. functional, task number | $|\rho|$ | PD median (std) | H median (std) |
|---|---|---|---|
| stroke speed, std, task 8 | 0.39 | 0.45 (0.88) | -0.47 (0.97) |
| relative NCP, task 8 | 0.37 | -0.22 (0.63) | -0.06 (1.16) |
| horizontal velocity, std , task 8 | 0.35 | 0.20 (0.99) | -0.33 (0.89) |
| $\rho_{vel}$, 99th percentile, task 2 | 0.35 | -0.57 (0.86) | 0.37(1.02) |
| horizontal velocity, 99th percentile, task 8 | 0.33 | 0.4 (0.95) | -0.44 (0.95) |
| $R^{fall}_{time}$, median, task 8 | 0.33 | -0.42 (0.57) | 0.11(1.22) |
| $\rho^{rise}_{horizontal,acc}$ , 1st percentile , task 8 | 0.33 | -0.17 (0.83) | -0.37 (1.06) |
| relative NCP, −, task 6 | 0.33 | -0.40 (0.56) | -0.1 (1.23) |
| $R^{rise}_{press}$, 99th percentile - 1st percentile, task 3 | 0.33 | -0.33(0.61) | 0.1 (1.2) |
| $\rho^{main}_{vertical,vel}$ , 99th percentile - 1st percentile , task 2 | 0.32 | -0.47 (0.75) | -0.11(1.12) |
| $R^{rise}_{press}$, std, task 3 | 0.32 | -0.29(0.6) | 0.06 (1.21) |
| horizontal jerk, std, task 8 | 0.32 | 0.23 (1.0) | -0.41 (0.9) |
| horizontal velocity , 99th percentile 99 - 1st percentile , task 8 | 0.32 | 0.22 (1.01) | -0.48 (0.89) |
| horizontal jerk, 99th percentile , task 8 | 0.32 | 0.29 (0.99) | -0.25 (0.91) |
| $R^{fall}_{time}$, median , task 3 | 0.31 | -0.17(0.65) | -0.1 (1.19) |
| velocity, median, task 8 | 0.31 | 0.2(0.9) | -0.35 (1.0) |
| horizontal velocity(rising edge) , std, task 8 | | 0.26 (0.98) | -0.44(0.94) |
| horizontal jerk, percentile 99th - percentile 1st, task 8 | 0.31 | 0.26 (0.97) | -0.33 (0.95) |
| $\rho_{vertical,vel}$, std, task 3 | 0.3 | -0.6 (0.93) | 0.23 (0.99) |
| velocity (rising edge), mean, task 8 | 0.3 | 0.16 (0.98) | -0.41 (0.93) |

Table 4: Classification accuracies of different handwriting tasks for kinematic and pressure features (support vector machines classifier).

| Evaluated task | $P_{acc}$ pressure features | $P_{acc}$ kinematic features | $P_{acc}$ kinematic and pressure |
|---|---|---|---|
| 1 (Archimedean spiral) | 62.8 | - | 62.8 |
| 2 (letter l) | 72.1 | 69.2 | 72.3 |
| 3 (bigram le) | 71.5 | 72.5 | 71.0 |
| 4 (word les) | 66.4 | - | 66.4 |
| 5 (word lektorka) | 66.9 | 65.1 | 65.2 |
| 6 (word porovnat) | 74.2 | 64.9 | 73.3 |
| 7 (word nepopadnout) | 66.8 | 66.4 | 67.6 |
| 8 (sentence) | 73.2 | 74.9 | 76.5 |
| overall | 82.5 | 75.4 | 81.3 |

Table 5: Comparison of different classifiers for diagnosis of Parkinson's disease from handwriting. Kinematic and pressure features obtained from tasks 2 − 8 were used.

| classifier | $P_{acc}$ [%] | $P_{spe}$ [%] | $P_{sen}$[%] |
|---|---|---|---|
| SVM | 81.3 | 80.9 | 87.4 |
| AdaBoost | 78.9 | 79.2 | 82.4 |
| $K$-NN | 71.7 | 70.8 | 78.5 |

## 4. Discussion

PD is a very complex disease with different symptoms that can vary from patient to patient. The handwriting process is a complex motor activity requiring the coordination of several muscles. Both these aspects make it very difficult to explain or exactly link any handwriting characteristics or features to particular symptoms of PD. The results of our study show that pressure or kinematic features can be used to support a differential diagnosis of PD; however, the exact relationship between PD symptoms and particular features is not known. From a clinical point of view, kinematic features reflect complex cognitive processes and are influenced by a wide range of clinical aspects such as tremor, muscle stiffness, rigidity, and variance in movement speed. On the other hand, the pressure features can provide additional information that is not captured in kinematic features.

As indicated in Table 4, not all handwriting tasks provide the same level of discrimination power. After evaluating our results, it is evident that some tasks are more useful for diagnosis than others. Task 8 appeared to be the most promising task. This is probably because the task involves writing a whole sentence, and some representations of PD appear only when the task has some temporal extension. This is similar to typical symptom of PD – micrography, where the letter size is reduced as the subject spends more time writing a sentence line. As in task 8, task 2 (writing the letter l) provided good predictive performance from kinematic and pressure features. There was a gap in the predictive performance derived from pressure features and kinematic features for the task 6 (writing the word nepopadnout). Tasks 1, 4, 5, and 7 did not contribute to overall predictive performance significantly, as they reached only 62 % – 66 % accuracy. These findings indicate that special attention should be paid by researchers and clinicians when designing handwriting templates or even handwriting standardized tests since the task selection strongly influences the results or the potential of acquired data. In this study we focus mainly on handwriting analysis and we did not utilise any spiral specific features. This may explain why the task 1 does not have a significant impact on classification. Therefore in our future work we plan to perform deeper spiral analysis and evaluate spiral specific features.

Decision support tools are gaining significant research interest due to their potential to improve health-care provision [5, 6]. Among many possible approaches, those that provide noninvasive monitoring and diagnosis of diseases are of increased interest to clinicians and biomedical engineers. We contribute to this area with the publication of our *PaHaW* database[3] , containing eight different handwriting samples from 75 healthy and PD subjects.

---

[3]The database can be downloaded from BDALab webpage (http://bdalab.utko.feec.vutbr.cz/) or UCI Machine Learning Repository (http://archive.ics.uci.edu/ml/)

To prove the relevance of the database, we proposed a methodology to build a predictive model of PD from kinematic and pressure handwriting measures. It was shown that using of basic kinematic and pressure features allowed for a classification accuracy of 82%. The proposed approach is not intended to replace the clinician but rather to provide assistance for a more accurate and objective diagnosis. When employed with other approaches such as speech processing [22, 31], even better results can be achieved in terms of accuracy of prediction. We showed that both kinematic and pressure features contribute in discriminating between PD and healthy subjects.

In this study, we almost 200 features. To follow and analyse such a high number of features can be very difficult for clinicians. It would be more convenient to specify a smaller representative group of features that would make it possible to map features to standard metrics and provide quantification of a PD severity. However this again requires a relatively high number of subjects with different levels of disease symptom severity.

The results presented in this study indicate that different aspects of handwriting can be with advantage used in diagnosis of PD. However, several limitations have to be recognized when interpreting the results. Firstly, in this study we decided to focus only on PD and healthy control group. Other diseases have been analysed in other papers [32, 33]. Inclusion of the cognitively well characterized PD patients at different motor stages of PD as well as inclusion of other relevant patient groups that suffer from both micrographia and cognitive impairment such as progressive supranuclear palsy or Huntington's disease are warranted in order to investigate whether the proposed technique can be used to discriminate between PD and other diseases. Classification of different diseases may be possible if they alter handwriting in diverse way, i.e. there are different patterns across the variables. However, this further underlines the importance of deeper handwriting analysis that include all modalities, since discriminative features may be hidden in handwriting signal and simple evaluation of conventional kinematic features may not be sufficient. Secondly, all patients with PD performed the handwriting tasks under medication. It suggests that proposed methodology may be sensitive enough to identify PD even if the symptoms are attenuated by the medication. On the other hand, medication can have side effects impacting the movements of the patients that can influence classification process. Before implementation of the proposed approach in the clinical settings a future study on patients without medication should be performed to investigate how would classifier perform under this condition. Thirdly, we have shown that handwriting can be used as biomarker for PD, however this should be considered only as a first step in further investigation. Longitudinal study is required to investigate subjects with a high risk of PD development, and confirm whether proposed methodology successfully identified participants that actually developed PD. This can reveal whether it is possible to use handwriting markers for early diagnosis of PD. Similarly, repeated measurements from the same subjects can be obtained to increase test-retest reliability or to investigate how can be proposed approach used for symptoms severity monitoring. Additionally, it is not possible to control for all intentional and unintentional alterations of handwriting. There may be many factors influencing handwriting that might impact classification decision. Therefore further studies are needed to provide confirmation of conclusions drawn in this paper. Standardized data collection and testing of subjects on multiple occasions is necessary and it is aim of our future research.

## Acknowledgements

This work was supported by the projects COST IC 1206, NT13499 (Speech, its impairment and cognitive performance in Parkinsons disease), LO1401, CZ.1.07/2.3.00/20.0094, project CEITEC, Central European Institute of Technology: (CZ.1.05/1.1.00/02.0068) from the European Regional Development Fund and by FEDER and Ministerio de Economa y Competitividad TEC2012-38630-C04-03. The described research was performed in laboratories supported by the SIX project; the registration number CZ.1.05/2.1.00/03.0072, the operational program Research and Development for Innovation. Peter Drotár was supported by the CZ.1.07/2.3.00/30.0005 project of Brno University of Technology.

## Appendix A.  Subject data

| ID | sex | diagnosis | age [years] | LED | UPDRS (part V) | Years since diag. |
|----|-----|-----------|-------------|------|----------------|-------------------|
| 01 | F | PD | 68 | 1115 | 2 | 6 |
| 02 | F | PD | 78 | 2110 | 2 | 8 |
| 03 | F | PD | 69 | 1557 | 2 | 7 |
| 04 | F | PD | 79 | 1691 | 2 | 12 |
| 05 | F | PD | 69 | 600 | 2 | 2 |
| 06 | F | PD | 57 | 1272 | 2 | 9 |

| 07 | F | PD | 78 | 666 | 3 | 19 |
|----|---|----|----|-----|---|----|
| 08 | F | PD | 58 | 397 | 1 | 5 |
| 09 | M | PD | 78 | 2066 | 1 | 3 |
| 10 | M | PD | 74 | 1480 | 2.5 | 3 |
| 13 | M | PD | 65 | 990 | 1 | 2 |
| 14 | M | PD | 64 | 1253 | 3 | 8 |
| 15 | F | PD | 69 | 990 | 2.5 | 17 |
| 16 | M | PD | 67 | 1188 | 2 | 4 |
| 17 | F | PD | 75 | 1370 | 5 | 18 |
| 18 | F | PD | 76 | 1250 | 2.5 | 17 |
| 19 | F | PD | 86 | 750 | 2 | 6 |
| 20 | F | PD | 79 | 2227 | 2 | 8 |
| 22 | F | PD | 67 | 645 | 2 | 14 |
| 23 | F | PD | 73 | 1235 | 2 | 9 |
| 24 | M | PD | 70 | 1317 | 4 | 7 |
| 25 | M | PD | 60 | 1143 | 3 | 10 |
| 26 | F | healthy | 57 | - | - | - |
| 27 | M | healthy | 92 | - | - | - |
| 28 | F | healthy | 52 | - | - | - |
| 29 | F | healthy | 58 | - | - | - |
| 30 | M | healthy | 69 | - | - | - |
| 31 | M | healthy | 76 | - | - | - |
| 32 | F | healthy | 59 | - | - | - |
| 33 | F | PD | 62 | 750 | 2 | 4 |
| 34 | M | PD | 61 | 2547 | 2 | 5 |
| 36 | M | PD | 90 | 750 | 2 | 3 |
| 39 | M | healthy | 65 | - | - | - |
| 40 | M | healthy | 53 | - | - | - |
| 41 | M | healthy | 78 | - | - | - |
| 43 | M | PD | 48 | 1080 | 1 | 4 |
| 44 | F | PD | 62 | 397 | 1 | 5 |
| 48 | M | PD | 87 | 1450 | 4 | 12 |
| 49 | M | healthy | 58 | - | - | - |
| 51 | F | healthy | 48 | - | - | - |
| 52 | F | healthy | 44 | - | - | - |
| 53 | M | PD | 84 | 1942 | 2 | 2 |
| 54 | M | PD | 69 | 2546 | 2 | 10 |
| 55 | M | PD | 63 | 1930 | 2.5 | 14 |

| 57 | M | healthy | 80 | - | - | - |
|----|---|---------|----|----|----|----|
| 60 | M | healthy | 65 | - | - | - |
| 61 | F | healthy | 59 | - | - | - |
| 62 | F | healthy | 65 | - | - | - |
| 66 | F | healthy | 69 | - | - | - |
| 67 | M | healthy | 59 | - | - | - |
| 69 | F | healthy | 74 | - | - | - |
| 70 | F | healthy | 47 | - | - | - |
| 71 | M | healthy | 52 | - | - | - |
| 72 | M | healthy | 45 | - | - | - |
| 73 | F | healthy | 64 | - | - | - |
| 74 | M | PD | 53 | 2387 | 2.5 | 9 |
| 75 | M | PD | 73 | 2010 | 2.5 | 12 |
| 76 | M | healthy | 56 | - | - | - |
| 77 | M | PD | 74 | 2337 | 3 | 1 |
| 78 | M | PD | 36 | 800 | 2 | 2 |
| 80 | M | PD | 67 | 3544 | 3 | 5 |
| 82 | M | healthy | 45 | - | - | - |
| 83 | F | healthy | 74 | - | - | - |
| 84 | F | healthy | 62 | - | - | - |
| 85 | F | healthy | 75 | - | - | - |
| 87 | M | healthy | 57 | - | - | - |
| 89 | M | healthy | 63 | - | - | - |
| 90 | M | healthy | 71 | - | - | - |
| 91 | F | healthy | 64 | - | - | - |
| 92 | F | healthy | 58 | - | - | - |
| 94 | M | healthy | 64 | - | - | - |
| 95 | M | healthy | 74 | - | - | - |
| 96 | F | healthy | 77 | - | - | - |
| 97 | M | healthy | 44 | - | - | - |
| 98 | F | PD | 77 | 1210 | 2 | 6 |

Table A.6: Detailed clinical and demographic information about participants.